\documentclass{relaxed_system_lab}

\usepackage[utf8]{inputenc}
\usepackage[T1]{fontenc}
\usepackage{geometry}
\usepackage{amsmath,amssymb}
\usepackage{courier}

\usepackage[toc,page,header]{appendix}

\usepackage{algorithm}
\usepackage[noend]{algorithmic}
\usepackage{array}
\usepackage{wrapfig}

\usepackage{amsmath}
\usepackage[utf8]{inputenc}
\usepackage[T1]{fontenc}
\usepackage{hyperref}
\usepackage{url}
\usepackage{booktabs}
\usepackage{amsfonts}
\usepackage[most]{tcolorbox}
\usepackage{nicefrac}
\usepackage[expansion=false]{microtype}
\usepackage{xcolor} 
\usepackage{amssymb}
\usepackage{caption}
\usepackage{xspace}
\usepackage{multirow}
\usepackage{graphicx}
\usepackage{subcaption}
\usepackage{enumitem}
\usepackage{pifont}
\usepackage{mathtools}
\usepackage{mathrsfs}
\usepackage{mdframed}
\usepackage{amsthm}
\usepackage[numbers]{natbib}

\definecolor{promptboxlightgray}{gray}{0.95}
\definecolor{promptboxblue}{rgb}{0.9, 0.9, 1.0}

\newcommand{\cmark}{\ding{51}}
\newcommand{\xmark}{\ding{55}}

\newtcolorbox{promptbox}[1][]{
  enhanced,
  breakable,
  colback=promptboxlightgray,
  colframe=promptboxblue!30,
  arc=8pt,
  boxrule=0.5pt,
  left=12pt,
  right=12pt,
  top=8pt,
  bottom=8pt,
  fonttitle=\bfseries,
  fontupper=\linespread{1.2}\selectfont,
  title=#1
}

\theoremstyle{plain}
\newtheorem{theorem}{Theorem}[section]
\newtheorem{proposition}[theorem]{Proposition}
\newtheorem{lemma}[theorem]{Lemma}
\newtheorem{corollary}[theorem]{Corollary}
\theoremstyle{definition}

\newtheorem{assumption}[theorem]{Assumption}
\theoremstyle{remark}

\newtheorem{example}{Example}

\title{Implicit Strategic Optimization: Rethinking Long-Horizon Decision-Making in Adversarial Poker Environments}

\author{Boyang Xia$^*$, Weiyou Tian$^*$, Qingnan Ren, Jiaqi Huang, Jie Xiao, \\ Shuo Lu, Kai Wang, Lynn Ai, Eric Yang, Bill Shi$^\dagger$}

\affiliation{Gradient}

\contribution{$^*$Equal contribution, $^\dagger$Corresponding author: \texttt{tianyu@gradient.network}}

\abstract{
Training large language model (LLM) agents for adversarial games is often driven by episodic objectives such as win rate. In long-horizon settings, however, payoffs are shaped by latent strategic externalities that evolve over time, so myopic optimization and variation-based regret analyses can become vacuous even when the dynamics are predictable. To solve this problem, we introduce \emph{Implicit Strategic Optimization} (ISO), a prediction-aware framework in which each agent forecasts the current strategic context and uses it to update its policy online. ISO combines a \emph{Strategic Reward Model} (SRM) that estimates the long-run strategic value of actions with \textsc{iso-grpo}, a context-conditioned optimistic learning rule. We prove sublinear contextual regret and equilibrium convergence guarantees whose dominant terms scale with the number of context mispredictions; when prediction errors are bounded, our bounds recover the static-game rates obtained when strategic externalities are known. Experiments in 6-player No-Limit Texas Hold'em and competitive Pok\'emon show consistent improvements in long-term return over strong LLM and RL baselines, and graceful degradation under controlled prediction noise.
}

\begin{document}

\maketitle

\section{Introduction}

Training large language model (LLM) agents for adversarial games is a natural testbed for sequential decision-making and strategic interaction, bridging classic online learning/game dynamics and modern agent training \cite{fudenberg1998theory,cesabianchi2006prediction}. In repeated games, multiplicative-weights and regret-based dynamics yield uncoupled learning with guarantees, and empirical play converges to equilibrium notions such as (coarse) correlated equilibrium \cite{freund1999adaptive,foster1997calibrated,hart2000simple,blum2007external}. Optimistic and regularized variants further sharpen convergence in games and saddle-point problems, with refined guarantees toward Nash equilibrium in structured settings \cite{rakhlin2013predictable,syrgkanis2015fast,daskalakis2019lastiterate,anagnostides2022omd}.

Many strategic environments are \emph{long-horizon}: action payoffs are mediated by latent strategic externalities (e.g., reputation, opponent adaptation, meta-game effects) that unfold over time and are not directly observable. Learning in \emph{time-varying} games is often analyzed via dynamic regret/tracking bounds scaling with path-length or variation \cite{zhang2022timevarying,anagnostides2023convergence,zinkevich2003online,chiang2012online,hall2013tracking,besbes2015nonstationary}. Yet variation can be large even when dynamics are highly predictable; optimistic analyses show that accurate forecasts can drastically reduce regret \cite{rakhlin2013predictable,chiang2012online}. As in Example~1, existing bounds may still grow linearly with the horizon even under deterministic strategic externalities, where constant regret is achievable—suggesting that agents must \emph{forecast future strategic returns}.

In parallel, large-scale systems achieve strong equilibrium-approximating behavior in long-horizon, imperfect-information games, including poker agents (DeepStack, Libratus, Pluribus) \cite{moravvcik2017deepstack,brown2019pluribus}, DeepNash for Stratego \cite{perolat2022mastering}, and population-based training such as AlphaStar \cite{vinyals2019grandmaster}. LLM-centric agents further combine language models with planning/search/opponent modeling (e.g., Cicero, Pok\'eChamp, LATS), and benchmarks like GameBench probe strategic reasoning across games \cite{bakhtin2022cicero,karten2025pokechamp,zhou2023lats,costarelli2024gamebench}. Despite empirical success, most methods optimize episodic objectives or assume stationary payoffs, leaving limited guarantees that tie \emph{prediction accuracy under evolving strategic externalities} to regret and equilibrium convergence.

Motivated by the central role of predictive models in economic and industrial decision-making \cite{jordan2015mltrends,varian2014bigdata,mullainathan2017mlapplied,athey2017beyond}, we ask:
\emph{How does an agent’s ability to accurately predict future strategic returns affect long-run performance and equilibrium convergence in long-horizon games?}
We take a \emph{prediction-aware} perspective and introduce \emph{Implicit Strategic Optimization} (ISO), which makes explicit what agents can forecast about latent strategic externalities and how such forecasts shape learning dynamics. Our main contributions are:
\begin{itemize}
    \item \textbf{Prediction-aware game framework (ISO).}
    We formalize long-horizon games with a latent context driving strategic externalities: players receive private context predictions, play mixed strategies conditioned on predictions, and update both policies and predictors online. We define contextual regret benchmarks and context-conditioned equilibrium notions.
    
    \item \textbf{Algorithm and theory with explicit dependence on mispredictions.}
    We propose \textbf{\textsc{iso-grpo}}, a context-conditioned optimistic learning rule. Via a contextual Regret bounded by Variation in Utility (RVU) analysis, we show regret and equilibrium convergence scale with the number of context mispredictions; under bounded prediction error, we recover static-game rates as if strategic externalities were known.
    
    \item \textbf{Empirical validation in complex games.}
    We evaluate ISO/\textsc{iso-grpo} in 6-player No-Limit Texas Hold'em and competitive Pok\'emon, showing improved long-term strategic returns over strong LLM and RL baselines and graceful degradation under controlled prediction noise.

\end{itemize}

\section{Related Works}

\subsection{GRPO for LLM Agents}

Reinforcement learning (RL) is widely used to train and align LLM agents under outcome-based or verifiable rewards, with PPO \cite{schulman2017proximal} as a common baseline but requiring a learned critic at substantial LLM-scale cost. Group Relative Policy Optimization (GRPO) \cite{shao2024deepseekmath} removes the critic by estimating advantages through group-relative reward comparisons, enabling scalable RL from verifiable rewards; subsequent work improves optimization/robustness in episodic regimes, including Token-Regulated GRPO \cite{le2025token}, Scaffolded GRPO \cite{zhang2025scaf}, and Training-Free GRPO \cite{cai2025training}. Existing GRPO-based methods, however, typically target short-horizon objectives and do not explicitly model evolving strategic context in long-horizon competitive interactions.

\subsection{Long-Horizon Adversarial Games and Strategic Learning for LLM Agents}

Strategic learning in adversarial multi-agent settings spans self-play in perfect-information games \cite{silver2018general} and equilibrium-seeking methods for imperfect-information domains such as poker, where RL and counterfactual regret minimization yield superhuman systems (e.g., Libratus, Pluribus) \cite{brown2018superhuman,brown2019pluribus}. Population-based game-theoretic training, including Policy Space Response Oracles (PSRO), further addresses non-stationary opponents and is supported by platforms like OpenSpiel \cite{lanctot2017unified,lanctot2019openspiel}, while recent work studies adversarial interaction directly among language agents \cite{cheng2024self}. Long-horizon LLM-agent training adds challenges from latent strategic externalities and opponent adaptation, motivating frameworks that combine contextual regret, optimism, and equilibrium convergence ideas \cite{freund1999adaptive,lattimore2020bandit}; our work contributes by extending GRPO toward prediction-aware strategic optimization in such long-horizon adversarial settings.

\begin{figure*}[h]
  \centering
  \includegraphics[width=\textwidth]{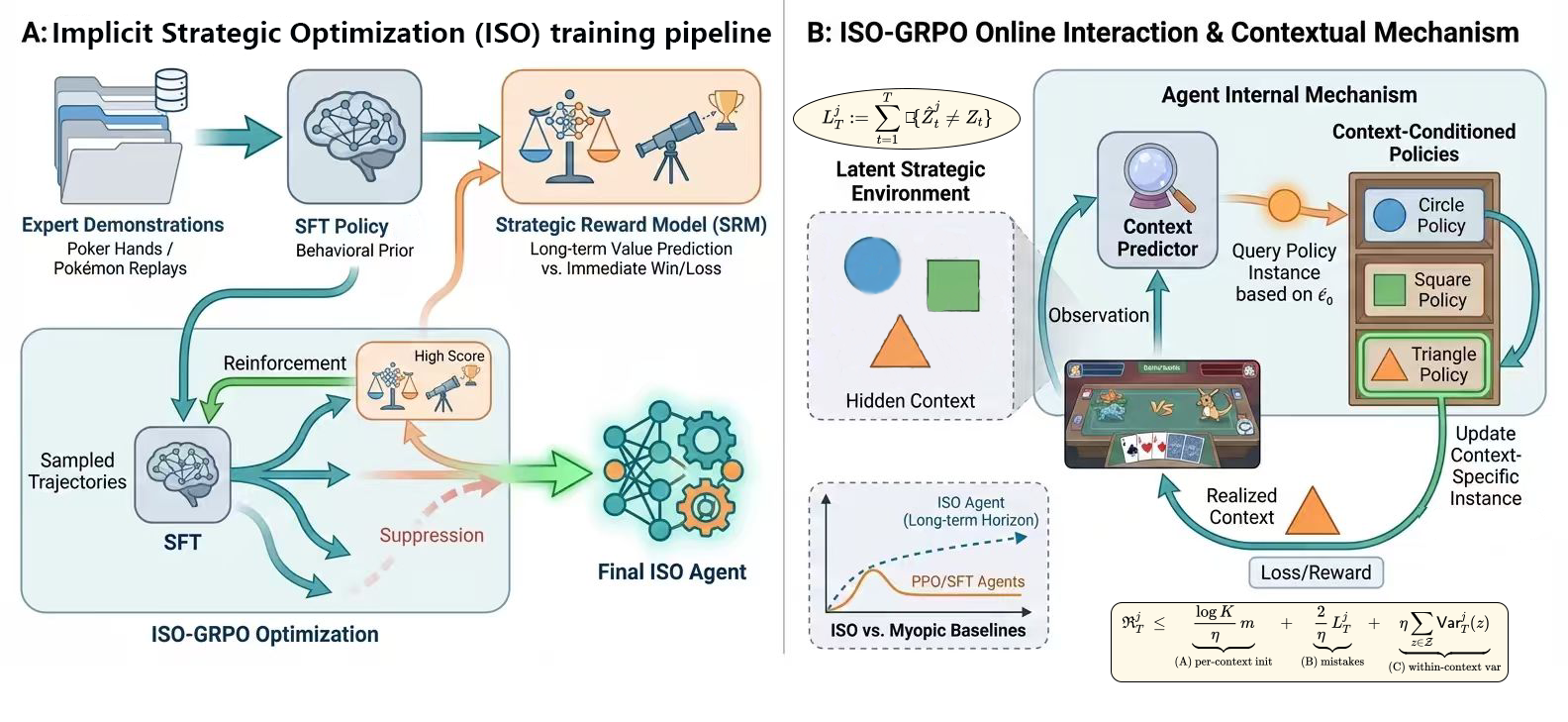}
  \caption{Overall ISO framework. (A) ISO training pipeline: expert demonstrations are distilled into an SFT policy, a SRM learns long-horizon strategic value, and ISO-GRPO optimizes the final agent using SRM-guided rewards. (B) ISO-GRPO online mechanism: the agent predicts latent strategic context, routes action selection through the predicted-context policy instance, observes realized context and reward, and updates the corresponding context-specific learner, yielding prediction-aware long-horizon adaptation.}
  \label{fig:iso_overview}
\end{figure*}

\section{Model}
\label{sec:model}
Figure~\ref{fig:iso_overview} summarizes the full ISO framework.
Panel (A) shows the offline training pipeline---expert demonstrations $\rightarrow$ SFT behavioral prior $\rightarrow$ Strategic Reward Model (SRM) $\rightarrow$ ISO-GRPO optimization.
Panel (B) depicts the online interaction mechanism in which the agent predicts a latent strategic context, queries a context-conditioned policy instance, receives realized context and reward, and updates the corresponding context-specific learner.
\paragraph{Notation.}
For an integer $n$, let $[n]=\{1,\dots,n\}$.
For a vector $x\in\mathbb{R}^d$, $x[\ell]$ denotes its $\ell$-th coordinate.
For a matrix $X\in\mathbb{R}^{d\times K}$, $X[:,k]\in\mathbb{R}^d$ denotes its $k$-th column.
We write $\langle x,y\rangle=x^\top y$ for the Euclidean inner product.
Let $\Delta_K=\{w\in\mathbb{R}^K: w[k]\ge 0,\ \sum_{k=1}^K w[k]=1\}$ denote the probability simplex.
For a finite action space $\mathcal{A}$, $\mathscr{P}(\mathcal{A})$ denotes the set of probability measures over $\mathcal{A}$.
When $\mathcal{A}=\mathcal{A}^1\times\cdots\times\mathcal{A}^J$ and players use independent mixed strategies $w^j\in\Delta_K$,
we write the induced product distribution as $\mathbf{w}=w^1\otimes\cdots\otimes w^J$.

\paragraph{Players and actions.}
We consider $J\ge 2$ players indexed by $[J]$.
Each player $j$ has a finite action set $\mathcal{A}^j=\{a_1^j,\dots,a_K^j\}$ with joint action space
$\mathcal{A}=\mathcal{A}^1\times\cdots\times\mathcal{A}^J$.

\paragraph{Latent strategic context (what it represents).}
At each round $t$, the interaction is governed by a \emph{latent strategic context} $Z_t\in\mathcal{Z}$.
Importantly, $Z_t$ is \emph{not} a physical state, not a Markov state, and not payoff noise.
It is a compact label for ``which strategic logic is currently in effect''---e.g., opponent adaptation mode,
reputation phase, meta-game regime---that
(i) affects payoffs, (ii) evolves over time, and (iii) is \emph{hidden but structured}.
The purpose of introducing $Z_t$ is to \emph{factor} long-horizon complexity into
\emph{prediction of the strategic regime} and \emph{learning conditioned on the regime}.

\begin{assumption}[Finite context set]
\label{ass:H2}
The context space is finite: $\mathcal{Z}=\{z_1,\dots,z_m\}$.
\end{assumption}

\paragraph{Why finite $\mathcal{Z}$.}
Assumption~\ref{ass:H2} is \emph{not} claiming the world is inherently discrete;
it asserts that, from a strategic viewpoint, there are finitely many distinguishable long-horizon regimes.
This assumption is theoretically central because it converts long-horizon difficulty into
\emph{online multiclass prediction complexity} (mistake bounds / regret for predicting $Z_t$).
We discuss relaxations to infinite/continuous contexts in Appendix~\ref{app:infinite-context}.

\paragraph{Linear-in-context loss model (analysis convenience).}
For each player $j$, let $\phi^j:\mathcal{A}\to\mathbb{R}^d$ be a feature map.
Given a joint mixed strategy $\mathbf{w}\in\mathscr{P}(\mathcal{A})$ and context $Z\in\mathcal{Z}$,
player $j$ incurs expected \emph{cost}
\begin{equation}
c^j(\mathbf{w},Z)
~:=~
\mathbb{E}_{\mathbf{a}\sim\mathbf{w}}\!\left[\left\langle \phi^j(\mathbf{a}),\,Z\right\rangle\right].
\label{eq:cost-def}
\end{equation}
This bilinear structure is not the core contribution; it is a clean modeling choice that
makes ``predicting context'' directly useful for ``predicting losses,'' and connects naturally
to optimistic online learning (Section~\ref{sec:prediction}).

\begin{assumption}[Bounded costs]
\label{ass:H1}
For any $j\in[J]$, $\mathbf{a}\in\mathcal{A}$, and $Z\in\mathcal{Z}$,
$\big|\langle \phi^j(\mathbf{a}), Z\rangle\big|\le 1$.
\end{assumption}

\paragraph{Per-action loss vectors under independent play.}
When $\mathbf{w}=w^j\otimes\mathbf{w}^{-j}$ is a product distribution, define
\begin{equation}
\Phi^j(\mathbf{w}^{-j})[:,k]
~:=~
\mathbb{E}_{\mathbf{a}^{-j}\sim\mathbf{w}^{-j}}\!\left[\phi^j(a_k^j,\mathbf{a}^{-j})\right].
\label{eq:Phi-def}
\end{equation}
Then the induced per-action loss vector for player $j$ at context $Z$ is
\begin{equation}
\ell^j(\mathbf{w}^{-j},Z)
~:=~
\Phi^j(\mathbf{w}^{-j})^\top Z
~\in~
\mathbb{R}^K.
\label{eq:ell-def}
\end{equation}

\begin{lemma}[Linearization]
\label{lem:linearize}
For any player $j$ and context $Z$,
\begin{equation}
c^j(\mathbf{w},Z)
~=~
\left\langle w^j,~\ell^j(\mathbf{w}^{-j},Z)\right\rangle.
\label{eq:linearize}
\end{equation}
\end{lemma}
\noindent
Lemma~\ref{lem:linearize} shows that, conditional on opponents' strategies and the realized context,
each player faces a standard full-information online learning problem over $\Delta_K$.
A short proof is deferred to Appendix~\ref{app:proof-linearize}.

\paragraph{Interaction protocol with private context predictions.}
The game unfolds over $T$ rounds. At each round $t$:
\begin{enumerate}[leftmargin=1.2em]
    \item Nature selects a context $Z_t\in\mathcal{Z}$.
    \item Each player $j$ observes a private \emph{prediction} $\hat{Z}_t^j\in\mathcal{Z}$
    (e.g., from an online multiclass predictor).
    \item Player $j$ selects a mixed action $w_t^j\in\Delta_K$ as a function of its history and $\hat{Z}_t^j$,
    inducing joint play $\mathbf{w}_t$.
    \item After play, player $j$ observes feedback sufficient to construct $\ell_t^j := \ell^j(\mathbf{w}_t^{-j},Z_t)$.
\end{enumerate}
The only ``long-horizon uncertainty'' we explicitly model is the \emph{hidden strategic regime} $Z_t$;
learning difficulty will be tied to how well players can predict it.

\paragraph{Prediction errors.}
We quantify prediction quality by the number of mispredictions
\begin{equation}
L_T^j
~:=~
\sum_{t=1}^T \mathbb{1}\!\left\{\hat{Z}_t^j \neq Z_t\right\}.
\label{eq:mistakes}
\end{equation}

\paragraph{Prediction-aware regret benchmark (not dynamic regret).}
Let $\mathscr{F}^z=\{t\in[T]: Z_t=z\}$ be the subsequence of rounds at which context $z$ is realized.
Define the \emph{best per-context static comparator}
\begin{equation}
\pi_\star^j(z)\in\arg\min_{w\in\Delta_K}\sum_{t\in\mathscr{F}^z}\left\langle w,\ell_t^j\right\rangle.
\label{eq:best-per-context}
\end{equation}
This comparator corresponds to an oracle that \emph{perfectly anticipates the realized strategic regime} $Z_t$
and plays the best fixed action distribution for each regime.
Our contextual external regret is
\begin{equation}
\mathfrak{R}_T^j
~:=~
\sum_{t=1}^T
\Big(
\left\langle w_t^j,\ell_t^j\right\rangle
-
\left\langle \pi_\star^j(Z_t),\ell_t^j\right\rangle
\Big).
\label{eq:ctx-regret}
\end{equation}
Crucially, \eqref{eq:ctx-regret} is \emph{not} dynamic regret: the comparator does not track an arbitrary sequence,
but only switches when the underlying strategic regime switches.
Thus $\mathfrak{R}_T^j$ measures the cost of \emph{imperfect strategic anticipation} (imperfect prediction of $Z_t$),
rather than penalizing all non-stationarity via variation/path-length terms.

\paragraph{Where a ``variation term'' can still matter.}
Our theory will separate (i) the cost of \emph{mispredictions} $L_T^j$ and (ii) the difficulty of learning \emph{within}
each realized context subsequence $\mathscr{F}^z$.
Large cross-context shifts do not directly harm the bound unless they induce mispredictions.
However, if losses vary adversarially \emph{within the same context} $z$, the within-context term can still dominate;
this is precisely what the contextual RVU analysis in Section~\ref{sec:prediction} controls.

\section{Prediction-aware Learning}
\label{sec:prediction}

\paragraph{Goal and decomposition.}
Given the model in Section~\ref{sec:model}, the remaining uncertainty is the latent context $Z_t$.
Players observe only predictions $\hat Z_t^j$, which may be wrong.
Our goal is to design an \emph{uncoupled} learning rule that
(i) guarantees no-regret with respect to the prediction-aware benchmark \eqref{eq:ctx-regret},
and (ii) makes the dependence on prediction quality explicit:
\emph{fewer context mistakes} $\Rightarrow$ \emph{easier long-horizon learning}.

\subsection{Algorithmic principle: ``one learner per context''}

\paragraph{Why separate learners by context.}
A central design choice is to maintain a separate online learner for each context $z\in\mathcal{Z}$.
Conceptually, each context defines a ``subgame'' (a regime-conditioned loss sequence).
If the prediction is correct, the player updates and plays in the correct subgame.
If the prediction is wrong, the player pays a localized ``mistake cost'' but does \emph{not}
pollute the learners associated with other contexts.
This realizes the intended factorization:long-horizon game $~\approx~$ (a small set of near-static problems) + (a small number of context mistakes).

\subsection{Algorithm: \textsc{iso-grpo} (context-conditioned optimistic updates)}

\paragraph{High-level description.}
For each context $z\in\mathcal{Z}$, player $j$ maintains an optimistic online learner
(e.g., optimistic MWU / GRPO-style update) producing a distribution $\rho_{t,z}^j\in\Delta_K$
and an optimistic prediction object $\Psi_{t,z}^j$.
At round $t$, the player routes \emph{play} through the learner indexed by the predicted context $\hat Z_t^j$,
and routes \emph{updates} through the learner indexed by the realized context $Z_t$.

\begin{algorithm}[t]
\caption{\textsc{iso-grpo} (player $j$)}
\label{alg:iso-grpo}
\begin{algorithmic}
\STATE Initialize $(\rho_{1,z}^j,\Psi_{1,z}^j)$ for all $z\in\mathcal{Z}$
\FOR{$t=1$ to $T$}
\STATE Observe prediction $\hat{Z}_t^j\in\mathcal{Z}$
\STATE Play $w_t^j \leftarrow \rho_{t,\hat{Z}_t^j}^j$
\STATE Observe realized context $Z_t$ and loss vector $\ell_t^j=\ell^j(\mathbf{w}_t^{-j},Z_t)$
\STATE Update $(\rho_{t+1,Z_t}^j,\Psi_{t+1,Z_t}^j)$ using $\ell_t^j$
\STATE Leave $(\rho_{t+1,z}^j,\Psi_{t+1,z}^j)=(\rho_{t,z}^j,\Psi_{t,z}^j)$ for all $z\neq Z_t$
\ENDFOR
\end{algorithmic}
\end{algorithm}

\paragraph{Instantiation.}
The analysis below only requires a standard optimistic stability inequality (an RVU-type one-step bound).
Thus \textsc{iso-grpo} can be instantiated with optimistic MWU in theory,
and with GRPO-style updates in our LLM training pipeline (Section~\ref{sec:experiments}).

\subsection{Contextual RVU: only count variation \emph{within} a context}

\paragraph{Why standard variation bounds are pessimistic.}
Classical dynamic-regret analyses bound regret by a path-length/variation term over the \emph{entire} horizon.
This can be overly pessimistic in long-horizon strategic settings:
large regime changes can be \emph{highly predictable} and should not be paid for if correctly anticipated.
Our key refinement is therefore:
\emph{compute variation only on the subsequences corresponding to the same realized context}.

\paragraph{Within-context variation proxy.}
For each context $z\in\mathcal{Z}$, let $\mathscr{F}^z=\{t: Z_t=z\}$ and let $t_z(1)<t_z(2)<\cdots$
be its ordered elements.
Define the within-context variation (one convenient proxy) as
\begin{equation}
\mathsf{Var}_T^j(z)
~:=~
\sum_{r\ge 2}
\left\|\ell_{t_z(r)}^j - \ell_{t_z(r-1)}^j\right\|_\infty.
\label{eq:within-var}
\end{equation}
This term ignores cross-context shifts entirely; it measures only how adversarially losses evolve \emph{inside}
the same strategic regime.

\begin{proposition}[Contextual regret bound (mispredictions drive regret)]
\label{prop:ctx-rvu}
Assume Assumptions~\ref{ass:H2}--\ref{ass:H1}.
Fix any player $j$ running \textsc{iso-grpo} with a stepsize $\eta\in(0,1]$ and an optimistic update rule
satisfying a standard RVU stability inequality within each context subsequence.\footnote{A sufficient condition and a concrete
instantiation with optimistic MWU are provided in Appendix~\ref{app:proof-ctx-rvu}.}
Then the contextual regret \eqref{eq:ctx-regret} satisfies
\begin{equation}
\label{eq:ctx-rvu-bound}
\begin{aligned}
\mathfrak{R}_T^j
~\le~&
\underbrace{\frac{\log K}{\eta}\,m}_{\text{(A) per-context init}}
+
\underbrace{\frac{2}{\eta}\,L_T^j}_{\text{(B) mistakes}}
+
\underbrace{\eta\sum_{z\in\mathcal{Z}}\mathsf{Var}_T^j(z)}_{\text{(C) within-context var}} \, .
\end{aligned}
\end{equation}
\end{proposition}

\paragraph{Interpretation of \eqref{eq:ctx-rvu-bound}.}
Term (B) is the key: regime shifts do not directly appear, only the number of times the player routes play using the wrong regime.
Thus:
predictable long-horizon (small $L_T^j$) \quad$\Rightarrow$\quad easy (near-static) learning.

Term (C) is the \emph{only} place where a ``variation'' notion can dominate:
if losses are adversarial \emph{within} the same context, regret can still be large.
This cleanly answers the ``corner case'' concern: prediction errors dominate when within-context sequences are stable/predictable,
but an adversary can still force difficulty by making losses vary rapidly without changing the context label.

\paragraph{Choosing $\eta$.}
A standard choice is $\eta \asymp \sqrt{\frac{m\log K + L_T^j}{\sum_z \mathsf{Var}_T^j(z)+1}}$,
yielding sublinear regret whenever $L_T^j=o(T)$ and the within-context variation grows sublinearly.

\subsection{Equilibrium convergence consequence}

\paragraph{From regret to equilibrium (coarse correlated equilibrium).}
When every player achieves sublinear contextual regret \eqref{eq:ctx-regret},
the empirical distribution of play converges to an approximate coarse correlated equilibrium (CCE)
in the standard way for uncoupled no-regret dynamics.
Concretely, letting $\bar{\mathbf{w}}_T:=\frac{1}{T}\sum_{t=1}^T \mathbf{w}_t$ denote the time-average joint distribution,
the CCE violation scales with the average regret.

\begin{corollary}[Approximate CCE with prediction dependence]
\label{cor:cce}
Under the assumptions of Proposition~\ref{prop:ctx-rvu}, if all players $j\in[J]$ run \textsc{iso-grpo}, then
$\bar{\mathbf{w}}_T$ is an $\varepsilon_T$-approximate CCE with
\begin{equation}
\varepsilon_T
~\le~
\frac{1}{T}\sum_{j=1}^J \mathfrak{R}_T^j,
\label{eq:cce-eps}
\end{equation}
and $\varepsilon_T\to 0$ whenever each $\mathfrak{R}_T^j=o(T)$.
In particular, via \eqref{eq:ctx-rvu-bound}, the dominant degradation from the long-horizon regime structure is controlled by
the cumulative mistake counts $L_T^j$ and the within-context terms $\sum_z \mathsf{Var}_T^j(z)$.
\end{corollary}

\section{Experiments}
\label{sec:experiments}
\subsection{Experimental Setup}
\label{subsec:exp_setup}
\paragraph{Tasks and environments.}
We evaluate \emph{Implicit Strategic Optimization} (ISO) in two long-horizon games:
(i) \textbf{6-player No-Limit Texas Hold'em (NLHE)} in a custom environment with heterogeneous agents (LLMs, RL, and rule-based bots). We use standard rules with \textbf{100BB} stacks and blinds \textbf{0.5BB/1BB}. Actions are \{\texttt{fold, check, call, bet, raise}\} with a continuous sizing component; for stability and reproducibility, we discretize bet/raise sizes into a small set of common sizes (e.g., pot fractions and all-in) while retaining strategic richness.
(ii) \textbf{competitive Pok\'{e}mon (Gen~1 OU)} via Pok\'{e}mon Showdown, using a fixed ladder-consistent ruleset (species clause) and a diverse opponent pool to reflect metagame uncertainty.

\paragraph{Observation and action interface.}
For LLM agents, we use a structured-text state encoding: (i) public state (street, pot, betting history, stacks), (ii) private information (hole cards; team/HP), and (iii) a constrained schema listing only legal actions and admissible moves. This standardizes prompts and makes comparisons across agents consistent.

\paragraph{Training pipeline.}
ISO-LLM follows three stages. \textbf{SFT:} we fine-tune \texttt{qwen2.5-0.5b} on curated expert gameplay (NLHE hand histories and solver-labeled actions at key subgames such as river; Pok\'{e}mon high-ELO replays and standard OU archetypes).
\textbf{SRM:} we train a strategic reward model to predict \emph{long-horizon strategic value} from trajectories, using delayed-return supervision so that locally suboptimal actions can be assigned high strategic value.
\textbf{GRPO:} starting from the SFT policy, we apply GRPO using SRM as the reward signal to optimize long-horizon value under diverse opponent interactions.

\paragraph{Baselines.}
We compare ISO-LLM with SFT-LLM, GPT-4o, Claude 3.5 Sonnet, Llama-3.1-70B, PPO, and CFR.

\paragraph{Metrics and diagnostics.}
We report Long-Term Return (LTR), Win Rate, Exploitability, and Strategic Sacrifice Rate (SSR). To localize gains, we additionally measure: (i) street-wise value decomposition, (ii) pot-size conditioned EV, (iii) negative-immediate / positive-long-horizon action rates (NLHE ``sacrifice''), and (iv) SRM calibration (rank correlation between SRM scores and realized long-horizon outcomes).

\paragraph{Evaluation protocol.}
Unless stated otherwise, NLHE results average over \textbf{10,000} hands with standard errors across multiple seeds and seating permutations. For Pok\'{e}mon, Elo is computed from large match sets against a mixed opponent pool and reported after rating convergence. Full details are in Appendix~\ref{app:exp}.

\subsection{Main Results}
\label{subsec:main_results}

We compare ISO-LLM against strong LLM, RL, and game-theoretic baselines in both environments.
Table~\ref{tab:main_results_6player} summarizes 6-player NLHE performance (10,000 hands), and Table~\ref{tab:pokemon_gpt4o} reports a cross-model Pok\'{e}mon evaluation against GPT-4o (for a Qwen3-1.7B backbone at different training stages).

\begin{table}[!htbp]
  \centering
  \caption{Main results in 6-player No-Limit Texas Hold'em. LTR denotes long-term return (BB/100 hands). Exploitability is measured in mBB per hand. Results are averaged over 10,000 hands.}
  \label{tab:main_results_6player}
  \small
  \setlength{\tabcolsep}{4pt}
  \renewcommand{\arraystretch}{1.2}
  \begin{tabular}{lccc}
    \toprule
    \textbf{Agent} & \textbf{LTR} & \textbf{Win Rate (\%)} & \textbf{Exploit. (mBB)} \\
    \midrule
    \textbf{ISO-LLM (Ours)}   & $+15.8$ & 22 & 5.2 \\
    SFT-LLM                  & $+3.1$  & 18 & 8.4 \\
    PPO                      & $+5.6$  & 25 & 7.1 \\
    CFR                      & $+1.2$  & 17 & 3.8 \\
    GPT-4o                   & $-8.2$  & 14 & 12.5 \\
    Claude 3.5 Sonnet        & $-5.4$  & 15 & 11.2 \\
    Llama-3.1-70B            & $-6.8$  & 13 & 13.8 \\
    \bottomrule
  \end{tabular}
\end{table}

\begin{table}[!htbp]
  \centering
  \caption{Cross-model evaluation in the Pok\'{e}mon environment against GPT-4o.
  We report the win rate of a Qwen3-1.7B agent at different training stages.}
  \label{tab:pokemon_gpt4o}
  \begin{tabular}{lc}
    \toprule
    \textbf{Model Variant} & \textbf{Win Rate (\%)} \\
    \midrule
    Origin (No FT) & 40 \\
    SFT            & 61 \\
    GRPO (ISO)     & \textbf{70} \\
    \bottomrule
  \end{tabular}
\end{table}

\begin{figure}[h]
    \centering
    \includegraphics[width=0.7\textwidth]{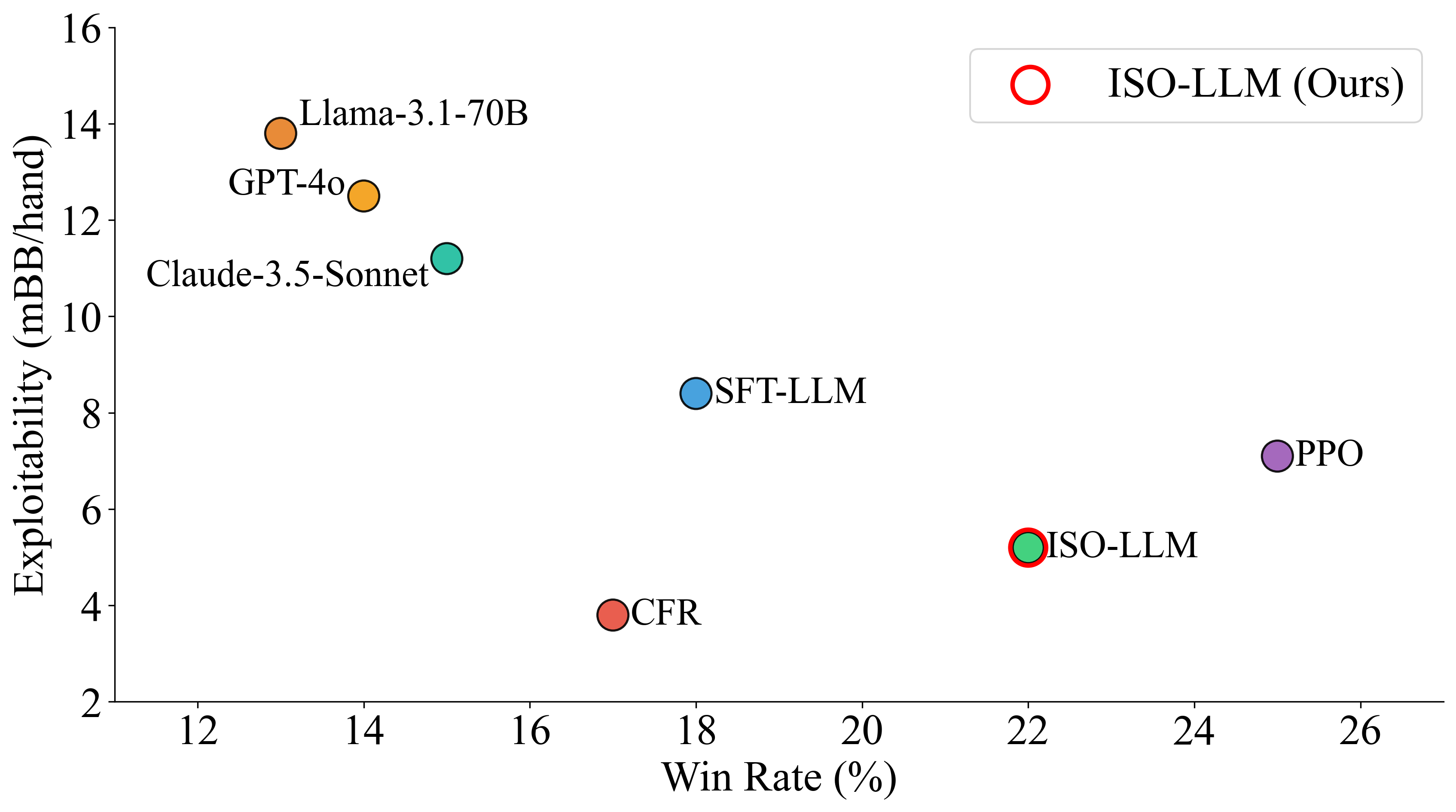}
    \caption{Win rate vs. exploitability trade-off.}
    \label{fig:poker_main}
\end{figure}

\paragraph{Analysis on NLHE (6-player).}
ISO-LLM achieves the best long-term return (LTR $+15.8$ BB/100) while remaining relatively hard to exploit (5.2 mBB/hand; Table~\ref{tab:main_results_6player} and Figure~\ref{fig:poker_main}).
Three takeaways are consistent with the ISO motivation:
(i) \textbf{win rate is not a proxy for long-horizon quality}: PPO attains the highest win rate (25\%) but substantially lower LTR, suggesting myopic optimization that wins many small pots while losing value in high-leverage spots;
(ii) \textbf{zero-shot LLMs are not robust} (negative LTR), indicating that language competence alone is insufficient without execution-grounded training;
(iii) CFR is the least exploitable among the listed methods but yields only modest LTR, consistent with equilibrium-oriented play being less effective against heterogeneous, non-equilibrium opponents in multi-player tables.

\paragraph{Analysis on Competitive Pok\'{e}mon (Gen~1 OU).}
ISO-LLM yields strong long-horizon performance (Elo 1750; win rate 68\%; SSR 15.2\%), consistent with delayed-payoff behaviors that trade short-term advantage for longer-term positional gain.
Table~\ref{tab:pokemon_gpt4o} isolates cross-model generalization: SFT provides a large jump (40$\rightarrow$61), and ISO-style GRPO yields a further gain to 70\% against GPT-4o.



\begin{figure}[htbp]
    \centering
    \includegraphics[width=0.9\textwidth]{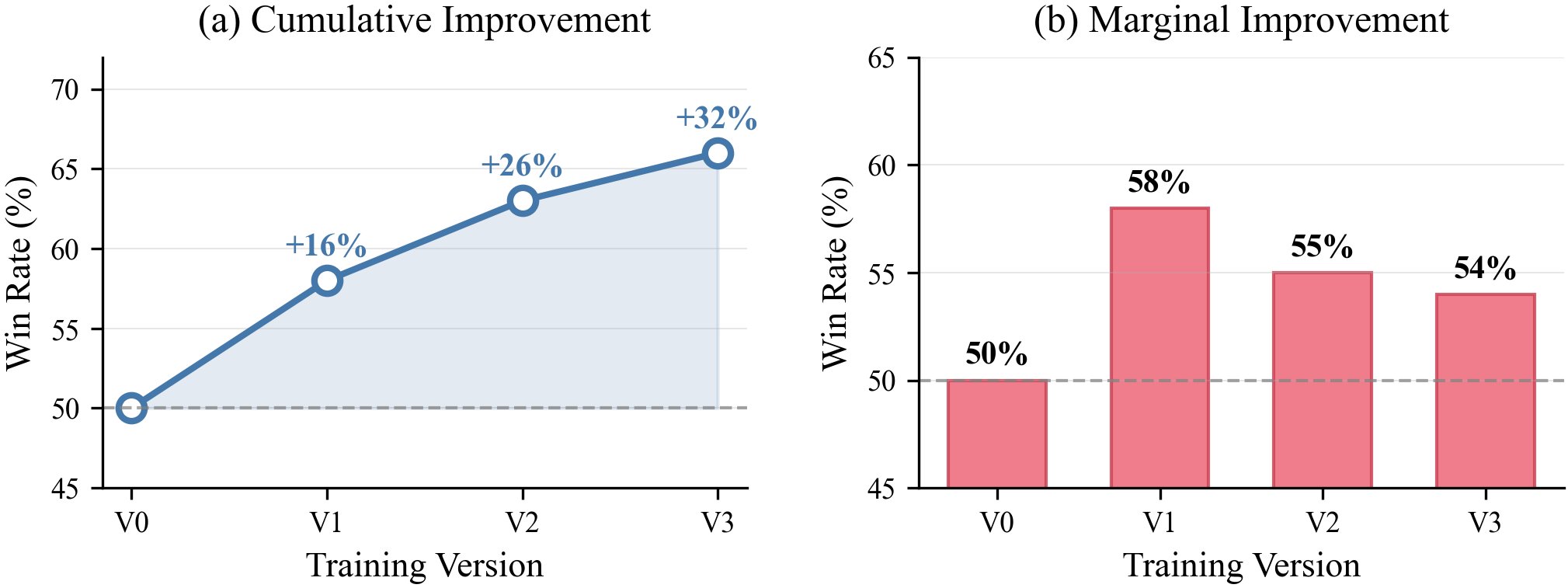}
    \caption{Iterative training improvement in Competitive Pokémon. (a) Cumulative improvement shows win rate against the baseline V0 across training versions. (b) Marginal improvement shows win rate against the immediate predecessor. ISO achieves consistent and monotonic gains.}
    \label{fig:pokemon_iterative}
\end{figure}

\paragraph{Iterative population-based improvement.}
We test training stability via population-based self-play over four versions (V0--V3), where each new version is trained against the full population of predecessors.
As shown in Figure~\ref{fig:pokemon_iterative}, ISO improves monotonically both against the baseline V0 (up to 66\% for V3) and against the immediate predecessor (positive margins of +8\%, +5\%, +4\%), suggesting stable incremental learning rather than oscillatory dynamics.

\begin{figure}[h]%
    \centering
    \includegraphics[width=0.65\linewidth]{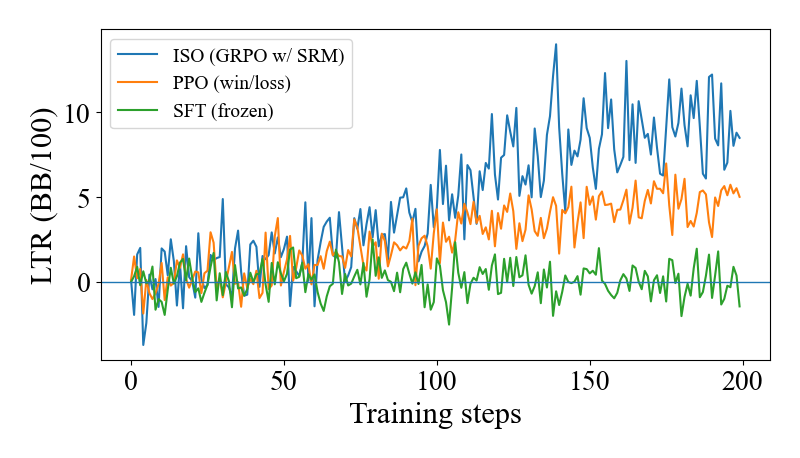}
    \caption{Training Curve of Poker}
    \label{fig:pokemon_training_curve}
\end{figure}

\paragraph{Training stability.}
Figure~\ref{fig:pokemon_training_curve} reports the GRPO reward trajectory: the smoothed reward increases over training with bounded variance, consistent with SRM providing a stable learning signal under the non-stationarity induced by population sampling.

\subsection{Ablation Study}
\label{subsec:ablation}

\paragraph{Isolating context-conditioned learning.}
We isolate ISO's algorithmic contribution beyond SRM and GRPO by varying how contextual predictions are used.
Table~\ref{tab:easy-ablation-context-components} shows that using predictions only as features yields limited gains, while context-separated learners with prediction-based routing and per-context updates markedly improve LTR and reduce exploitability.
Oracle routing with true $Z_t$ indicates the remaining gap is mostly due to prediction errors.

\begin{table}[!htbp]
  \centering
  \setlength{\tabcolsep}{4pt}
  \renewcommand{\arraystretch}{1.15}
  \caption{Ablation of context-conditioned components. \textbf{Sep}: per-context learners. \textbf{Route}: prediction used to route actions (feat = used as feature only; pred = routing + per-ctx update; oracle = true $Z_t$). More details are provided in Appendix~\ref{AS}.}
  \label{tab:easy-ablation-context-components}
  \begin{tabular}{lccrr}
    \toprule
    \textbf{Method} & \textbf{Sep} & \textbf{Route} & \textbf{LTR} & \textbf{Expl.} \\
    \midrule
    GRPO            & \xmark & --     & 2.7 & 7.6 \\
    GRPO + pred feat& \xmark & feat   & 3.1 & 7.1 \\
    \textbf{ISO-GRPO}& \cmark & pred   & \textbf{6.3} & \textbf{4.9} \\
    ISO-GRPO (oracle)& \cmark & oracle & 6.8 & 4.7 \\
    \bottomrule
  \end{tabular}
\end{table}

\begin{figure}[h]
    \centering
    \includegraphics[width=0.6\textwidth]{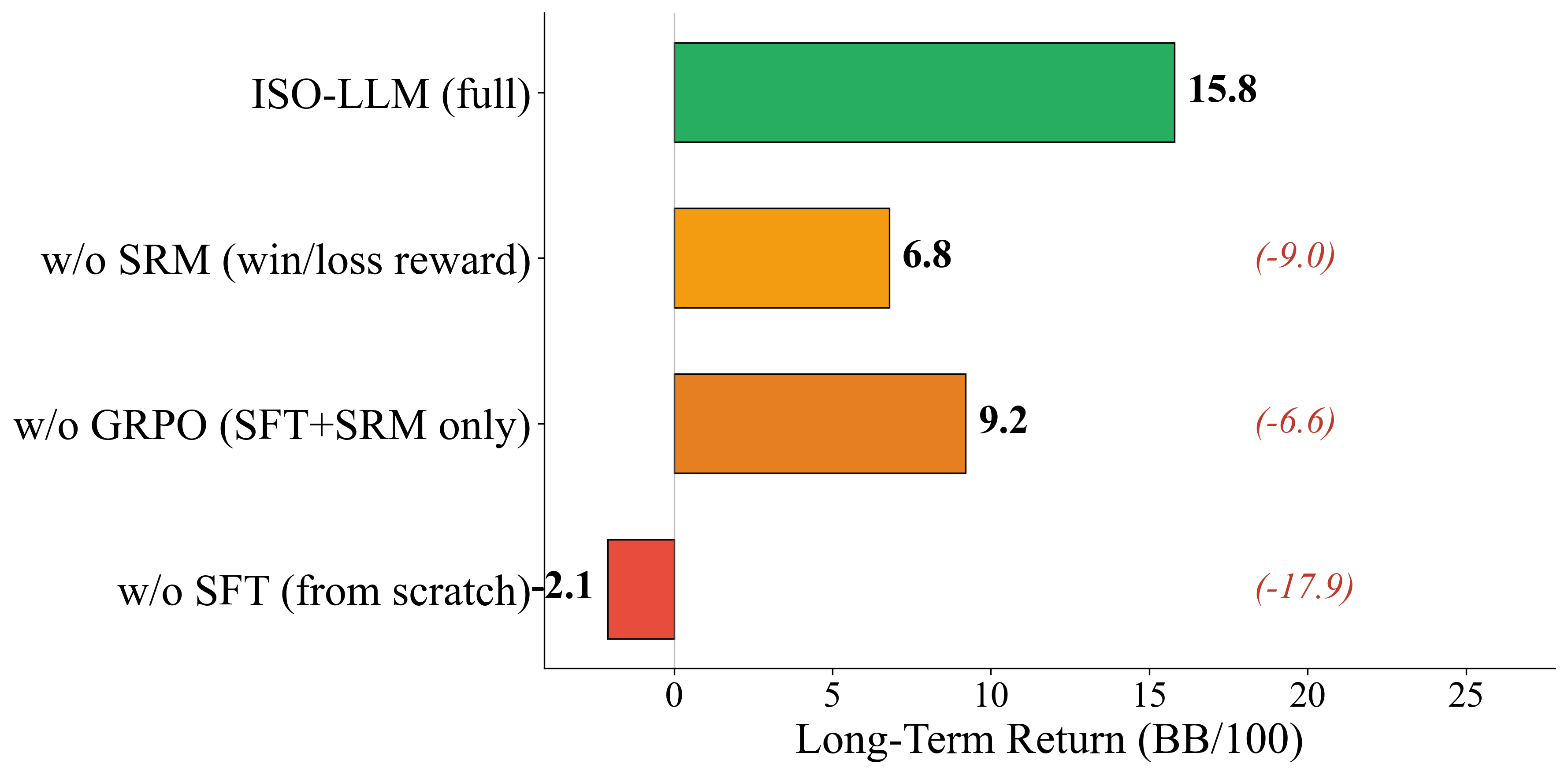}
    \caption{Ablation study of ISO components in 6-player NLHE. Removing SRM, GRPO, or SFT each leads to substantial performance degradation, indicating that all three components are necessary for strong long-horizon strategic performance.}
    \label{fig:ablation}
\end{figure}

To isolate the contribution of ISO components, we conduct ablations in NLHE (Figure~\ref{fig:ablation}).
Across variants, all three components are necessary for strong performance:
(i) replacing SRM with win/loss reward substantially reduces LTR, indicating that delayed-value supervision is essential;
(ii) removing GRPO reduces LTR, suggesting that optimization is necessary beyond reward modeling;
and (iii) removing SFT leads to negative LTR, highlighting the importance of a strong behavioral prior.

\paragraph{Prediction accuracy sweep.}
To directly align with our theory, we control the misprediction rate
$\epsilon = \Pr[\hat Z_t \neq Z_t]$, so that $L_T/T \approx \epsilon$.
Figure~X shows that ISO improves monotonically with prediction accuracy and
approaches the oracle-context upper bound as $\epsilon \to 0$.
In contrast, random-context routing does not benefit from increased predictability.

\subsection{Mechanistic Analysis}
\label{subsec:mechanistic}

We test whether ISO's gains reflect long-horizon competence rather than win-rate optimization.

\paragraph{Street-wise value decomposition.}
Table~\ref{tab:street_decomp} decomposes LTR by street and shows that ISO's gains concentrate on turn and river.

\begin{table}[h]
  \centering
  \caption{Street-wise value decomposition in 6-player NLHE.
  Entries are contributions to total LTR, measured in BB/100; rows sum to Table~1.}
  \label{tab:street_decomp}
  \small
  \renewcommand{\arraystretch}{1.1}
  \begin{tabular}{lccccc}
    \toprule
    \textbf{Agent} & \textbf{Pre} & \textbf{Flop} & \textbf{Turn} & \textbf{River} & \textbf{Total} \\
    \midrule
    \textbf{ISO-LLM} & $+2.1$ & $+3.4$ & \textbf{+4.6} & \textbf{+5.7} & \textbf{+15.8} \\
    SFT-LLM          & $+1.0$ & $+1.4$ & $+0.5$       & $+0.2$       & $+3.1$ \\
    PPO              & $+2.4$ & $+2.1$ & $+0.8$       & $+0.3$       & $+5.6$ \\
    CFR              & $+0.3$ & $+0.5$ & $+0.2$       & $+0.2$       & $+1.2$ \\
    GPT-4o           & $-1.0$ & $-2.0$ & $-2.4$       & $-2.8$       & $-8.2$ \\
    \bottomrule
  \end{tabular}
\end{table}

\paragraph{Pot-size conditioned EV.}
We bucket hands by terminal pot size and report EV (BB/hand).
Table~\ref{tab:pot-buckets} shows ISO remains positive in medium/large pots, whereas PPO degrades in large pots.

\begin{table}[h]
\caption{Pot-size conditioned EV in NLHE. EV is BB/hand within each terminal pot-size bucket.}
\label{tab:pot-buckets}
\centering
\setlength{\tabcolsep}{3pt}  
\renewcommand{\arraystretch}{1}  
\begin{tabular}{@{}lcccc@{}}
\toprule
Agent & $<5$BB & 5--20BB & 20--50BB & $>50$BB \\
\midrule
\textbf{ISO-LLM} & +0.010 & +0.018 & \textbf{+0.030} & \textbf{+0.060} \\
SFT-LLM & +0.006 & +0.008 & +0.004 & -0.005 \\
PPO & \textbf{+0.014} & +0.010 & -0.006 & -0.030 \\
CFR & +0.004 & +0.006 & +0.003 & +0.001 \\
GPT-4o & -0.005 & -0.012 & -0.020 & -0.040 \\
\bottomrule
\end{tabular}
\end{table}

\paragraph{Strategic sacrifice.}
We define a ``strategic sacrifice'' as an action with negative immediate EV under a shallow evaluator but positive empirical long-horizon return.
Table~\ref{tab:poker-ssr} shows ISO executes more such actions and converts them into larger LTR gains.

\begin{table}[h]
  \centering
  \caption{Poker strategic sacrifice analysis. NIE denotes actions with negative immediate EV under a shallow evaluator. 
  We report the NIE action rate, the fraction of such actions that yield positive long-horizon return, and the resulting LTR gain.}
  \label{tab:poker-ssr}
  \small
  \begin{tabular}{lrrr}
    \toprule
    \textbf{Agent} & \textbf{NIE Rate (\%)} & \textbf{LH+ Rate (\%)} & \textbf{LTR Gain} \\
    \midrule
    \textbf{ISO-LLM} & \textbf{12.8} & \textbf{61.0} & \textbf{+4.9} \\
    SFT-LLM          & 4.1           & 48.0          & +0.7 \\
    PPO              & 5.6           & 44.5          & +0.4 \\
    CFR              & 2.2           & 50.5          & +0.2 \\
    \bottomrule
  \end{tabular}
\end{table}

\subsection{Robustness: Opponent Populations and Prediction Noise}
\label{subsec:robustness}

We evaluate whether ISO generalizes across opponent mixtures and how it degrades under prediction noise.

\paragraph{Population robustness.}
Table~\ref{tab:population-robust} reports LTR and exploitability across heterogeneous opponent mixtures and randomized seating.
ISO achieves consistently positive LTR with relatively stable exploitability, suggesting reduced population overfitting.

\begin{table}[h]
  \centering
  \caption{Population robustness in 6-player NLHE. 
  ISO-LLM is evaluated against different opponent mixtures. LTR is measured in BB/100 and exploitability in mBB per hand.}
  \label{tab:population-robust}
  \small
  \begin{tabular}{lrr}
    \toprule
    \textbf{Opponents (5)} & \textbf{LTR} & \textbf{Exploit.} \\
    \midrule
    5$\times$PPO                         & +18.6 & 6.0 \\
    3$\times$PPO + 2$\times$SFT          & +16.2 & 5.8 \\
    2$\times$CFR + 3$\times$PPO          & +12.4 & 5.5 \\
    1$\times$GPT-4o + 2$\times$PPO + 2$\times$SFT & +14.1 & 5.7 \\
    5$\times$ISO (self-play)             & +1.5  & 4.9 \\
    \bottomrule
  \end{tabular}
\end{table}

\paragraph{Prediction noise sensitivity.}
We inject noise into the predicted strategic context (or SRM conditioning signal), corrupting it with probability $p$.
Table~\ref{tab:noise-sweep} shows smooth degradation as $p$ increases: LTR decreases while exploitability increases gradually, also with prediction-aware guarantees.

\begin{table}[h]
  \centering
  \caption{Sensitivity to prediction noise. We corrupt the predicted strategic context (or SRM conditioning) with probability $p$.}
  \label{tab:noise-sweep}
  \small
  \setlength{\tabcolsep}{5pt}
  \renewcommand{\arraystretch}{1.05}
  \scalebox{0.96}{
  \begin{tabular}{lrrrrr}
    \toprule
    Noise $p$ & 0\% & 10\% & 30\% & 50\% & 70\% \\
    \midrule
    ISO LTR (BB/100) & \textbf{15.8} & \textbf{13.9} & \textbf{10.4} & 6.7 & 2.1 \\
    ISO Exploit. (mBB/hand) & 5.2 & 5.6 & 6.8 & \textbf{8.9} & \textbf{12.5} \\
    \bottomrule
  \end{tabular}
  }
\end{table}

\subsection{SRM Quality and Calibration}
\label{subsec:srm_quality}

Since SRM is central to ISO, we evaluate its predictive quality on held-out trajectories.
Table~\ref{tab:srm-quality} reports rank correlation (Spearman $\rho$, Kendall $\tau$) between SRM scores and realized long-horizon returns, as well as top-decile precision.

\begin{table}[h]
  \centering
  \caption{SRM quality on held-out trajectories.}
  \label{tab:srm-quality}
  \small
  \begin{tabular}{lrrr}
    \toprule
    \textbf{Model} & $\boldsymbol{\rho}$ & $\boldsymbol{\tau}$ & \textbf{Top-10 Prec.} \\
    \midrule
    \textbf{ISO-SRM (ours)} & \textbf{0.46} & \textbf{0.33} & \textbf{71.0} \\
    Win/Loss Critic         & 0.18          & 0.12          & 55.0 \\
    Random                  & 0.00          & 0.00          & 10.0 \\
    \bottomrule
  \end{tabular}
\end{table}

\section{Conclusion}

Recent progress on uncoupled learning for long-horizon games enables principled modeling of non-stationary strategic interaction, but existing analyses largely treat non-stationarity as purely adversarial and ignore agents’ ability to \emph{predict} evolving strategic externalities. We introduced \emph{Implicit Strategic Optimization} (ISO), a prediction-aware framework that factorizes long-horizon difficulty into (i) forecasting a latent strategic context and (ii) learning conditioned on that context. We proposed \textsc{iso-grpo}, a context-conditioned optimistic update rule that routes actions by predicted context and updates by realized context, yielding regret and equilibrium-convergence guarantees whose dominant degradation scales with context mispredictions. Empirically, ISO improves long-term strategic return in 6-player No-Limit Texas Hold’em and competitive Pok\'emon, highlighting a practical decoupling between episodic win rate and long-run strategic value.

Looking ahead, three directions are especially promising: (1) weaken feedback assumptions (e.g., bandit/partial feedback) while retaining prediction-dependent guarantees; (2) extend ISO beyond finite context sets to continuous or large context spaces; and (3) study strategic settings where players’ predictions are jointly inferred or socially coupled, and design learning rules that account for this interaction.

\section*{Impact Statement}

This work aims to advance machine learning by developing prediction-aware learning mechanisms for long-horizon, multi-agent decision-making. By explicitly modeling and accounting for prediction uncertainty, the proposed system can support more robust reasoning and adaptation in complex decision environments where future conditions or strategic contexts must be estimated rather than observed.

Such capabilities may benefit the design of AI systems for planning, coordination, and decision support in dynamic and uncertain settings. We do not anticipate immediate negative societal impacts arising specifically from this work; as with general-purpose learning systems, responsible deployment and domain-specific oversight remain important considerations.

\bibliography{example_paper}
\bibliographystyle{plainnat}

\appendix
\onecolumn

\section{Additional Proofs for Sections~\ref{sec:model}--\ref{sec:prediction}}
\label{app:additional-proofs}

\subsection{Proof of Lemma~\ref{lem:linearize}}
\label{app:proof-linearize}

Fix player $j$ and write $\mathbf{w}=w^j\otimes \mathbf{w}^{-j}$.
By definition,
\begin{equation}
\label{eq:app-linearize-1}
c^j(\mathbf{w},Z)
=
\mathbb{E}_{a^j\sim w^j,\ \mathbf{a}^{-j}\sim \mathbf{w}^{-j}}
\!\left[\left\langle \phi^j(a^j,\mathbf{a}^{-j}),Z\right\rangle\right].
\end{equation}
Swap expectation and inner product and condition on $a^j=a_k^j$:
\begin{equation}
\label{eq:app-linearize-2}
c^j(\mathbf{w},Z)
=
\sum_{k=1}^K w^j[k]\,
\left\langle
\mathbb{E}_{\mathbf{a}^{-j}\sim \mathbf{w}^{-j}}\!\left[\phi^j(a_k^j,\mathbf{a}^{-j})\right],
Z
\right\rangle.
\end{equation}
Using the definition of $\Phi^j(\mathbf{w}^{-j})[:,k]$ from \eqref{eq:Phi-def}, this becomes
\begin{equation}
\label{eq:app-linearize-3}
\begin{aligned}
c^j(\mathbf{w},Z)
&=
\sum_{k=1}^K w^j[k]\,
\left\langle
\Phi^j(\mathbf{w}^{-j})[:,k],
Z
\right\rangle \\
&=
\left\langle w^j, \Phi^j(\mathbf{w}^{-j})^\top Z\right\rangle
=
\left\langle w^j,\ell^j(\mathbf{w}^{-j},Z)\right\rangle,
\end{aligned}
\end{equation}
which is exactly \eqref{eq:linearize}.

\subsection{Proof sketch for Proposition~\ref{prop:ctx-rvu} (Contextual RVU)}
\label{app:proof-ctx-rvu}

We outline a sufficient argument based on a standard optimistic MWU/RVU inequality;
the same structure applies to other optimistic updates.

\paragraph{Step 1: Decompose regret by realized contexts.}
Let $\mathscr{F}^z=\{t: Z_t=z\}$. Then
\begin{equation}
\label{eq:app-ctx-rvu-1}
\mathfrak{R}_T^j
=
\sum_{z\in\mathcal{Z}}
\sum_{t\in \mathscr{F}^z}
\Big(
\langle w_t^j,\ell_t^j\rangle
-
\langle \pi_\star^j(z),\ell_t^j\rangle
\Big).
\end{equation}
For each $z$, compare the realized actions $w_t^j$ to the best fixed comparator for that same subsequence.

\paragraph{Step 2: Separate ``correct routing'' rounds from mispredicted rounds.}
Let $\mathscr{G}^j:=\{t: \hat Z_t^j=Z_t\}$ be correctly predicted rounds and $\mathscr{B}^j:=\{t: \hat Z_t^j\neq Z_t\}$ be mistakes.
Then $|\mathscr{B}^j|=L_T^j$.
On correct rounds, play is produced by the learner corresponding to the realized context; on mistake rounds, play comes from an incorrect learner.

\paragraph{Step 3: RVU bound within each context subsequence.}
Fix $z\in\mathcal{Z}$ and restrict attention to rounds in $\mathscr{F}^z$.
On those rounds where the player routes correctly ($t\in \mathscr{F}^z\cap \mathscr{G}^j$),
the per-context optimistic update (e.g., optimistic MWU) yields a standard RVU inequality of the form
\begin{equation}
\label{eq:app-ctx-rvu-2}
\sum_{t\in \mathscr{F}^z\cap \mathscr{G}^j}
\Big(\langle w_t^j,\ell_t^j\rangle-\langle \pi_\star^j(z),\ell_t^j\rangle\Big)
\le
\frac{\log K}{\eta}
+
\eta\,\mathsf{Var}_T^j(z),
\end{equation}
where $\mathsf{Var}_T^j(z)$ is any admissible within-context variation proxy; \eqref{eq:within-var} is one such choice.

\paragraph{Step 4: Pay a uniform penalty for mistake rounds.}
On mistake rounds $t\in \mathscr{B}^j$, we upper bound the instantaneous regret crudely using bounded losses.
By Assumption~\ref{ass:H1}, each coordinate of $\ell_t^j$ lies in $[-1,1]$, hence
\begin{equation}
\label{eq:app-ctx-rvu-3}
\langle w_t^j,\ell_t^j\rangle-\langle \pi_\star^j(Z_t),\ell_t^j\rangle
\le 2.
\end{equation}
A standard way to integrate this into an RVU-style bound for the routed learner is to charge an additional
$\frac{2}{\eta}$ per mistake (equivalently, to treat each mistake as breaking the potential decrease argument).
Summing over $L_T^j$ mistakes yields the $\frac{2}{\eta}L_T^j$ term in \eqref{eq:ctx-rvu-bound}.

\paragraph{Step 5: Sum over contexts.}
Summing the per-context bounds over $z\in\mathcal{Z}$ gives
\begin{equation}
\label{eq:app-ctx-rvu-4}
\begin{aligned}
\mathfrak{R}_T^j
&\le
\frac{\log K}{\eta}\,m
+
\frac{2}{\eta}\,L_T^j
+
\eta\sum_{z\in\mathcal{Z}}\mathsf{Var}_T^j(z),
\end{aligned}
\end{equation}
which matches \eqref{eq:ctx-rvu-bound}.

\section{Extension to Infinite Context Sets}
\label{app:infinite-context}

\begin{assumption}[Additional regularity for infinite context sets]
\label{ass:5}

We include a short derivation to make the argument self-contained. The proof is
RVU-style, with the only substantive change being that all quantities are
conditioned on the realized context sequence.

\paragraph{Notation.}
At round $t$, the learner observes $c_t\in\mathcal{C}$ and samples an action
$a_t \sim \pi_t(\cdot \mid c_t)$. Let $r_t(a,c)\in[0,1]$ be the reward and
$a_t^\star(c)\in\arg\max_a r_t(a,c)$ be an optimal action under context $c$.
Define the (random) instantaneous regret
\begin{equation}
\Delta_t
\;:=\;
r_t(a_t^\star(c_t),c_t) - r_t(a_t,c_t),
\qquad
\mathcal{R}_T := \sum_{t=1}^T \mathbb{E}[\Delta_t].
\end{equation}
We write $\mathbb{E}_t[\cdot]$ for conditional expectation w.r.t.\ the history
$\mathcal{H}_{t-1} := (c_1,a_1,r_1,\ldots,c_{t-1},a_{t-1},r_{t-1})$.

\paragraph{A contextual RVU decomposition.}
Let $\hat r_t(a,c)$ be the model prediction used by the algorithm and define
the \emph{prediction error} (under the realized context) as
\begin{equation}
\varepsilon_t
\;:=\;
\hat r_t(a_t,c_t) - r_t(a_t,c_t),
\qquad
\varepsilon_t^\star
\;:=\;
\hat r_t(a_t^\star(c_t),c_t) - r_t(a_t^\star(c_t),c_t).
\end{equation}
Then by adding and subtracting $\hat r_t(\cdot,c_t)$ we obtain
\begin{align}
\Delta_t
&=
\underbrace{\bigl[\hat r_t(a_t^\star(c_t),c_t)-\hat r_t(a_t,c_t)\bigr]}_{\text{(I) optimistic gap}}
\;+\;
\underbrace{\bigl[\varepsilon_t - \varepsilon_t^\star\bigr]}_{\text{(II) prediction error}}.
\label{eq:rvu-decomp}
\end{align}
Taking $\mathbb{E}_t[\cdot]$ on both sides yields the same decomposition in
expectation, now conditioned on $\mathcal{H}_{t-1}$ (and hence on $c_t$).

\paragraph{Bounding the prediction error term.}
Assume the algorithm maintains a context-conditional error bound
\begin{equation}
\label{eq:ctx-error}
|\varepsilon_t| \le \beta_t(c_t),
\qquad
|\varepsilon_t^\star| \le \beta_t(c_t),
\end{equation}
for some predictable radius $\beta_t(\cdot)$ (e.g., derived from concentration
or a calibrated score model). Then
\begin{equation}
\mathbb{E}_t[\varepsilon_t - \varepsilon_t^\star]
\;\le\;
\mathbb{E}_t[|\varepsilon_t|+|\varepsilon_t^\star|]
\;\le\; 2\,\beta_t(c_t).
\label{eq:pred-term}
\end{equation}

\paragraph{Bounding the optimistic gap term.}
The RVU argument controls term (I) via a stability / one-step improvement
inequality induced by the update rule. Concretely, suppose the policy update
satisfies, for each realized context $c_t$,
\begin{equation}
\label{eq:stability}
\mathbb{E}_t\!\left[\hat r_t(a_t^\star(c_t),c_t)-\hat r_t(a_t,c_t)\right]
\;\le\;
\frac{1}{\eta}\Bigl(\Phi(\pi_t,\!c_t)-\Phi(\pi_{t+1},\!c_t)\Bigr)
\;+\;
\eta \,\mathbb{V}_t(c_t),
\end{equation}
where $\eta>0$ is a stepsize, $\Phi(\cdot,c)$ is a context-conditional potential
(e.g., negative entropy / Bregman divergence), and $\mathbb{V}_t(c)$ is the
context-conditional variance proxy that appears in RVU-style bounds.

\paragraph{Putting everything together.}
Combining \eqref{eq:rvu-decomp}, \eqref{eq:pred-term}, and \eqref{eq:stability}
and summing over $t=1,\ldots,T$ yields
\begin{align}
\mathcal{R}_T
&\le
\frac{1}{\eta}\sum_{t=1}^T
\mathbb{E}\Bigl[\Phi(\pi_t,c_t)-\Phi(\pi_{t+1},c_t)\Bigr]
\;+\;
\eta \sum_{t=1}^T \mathbb{E}\bigl[\mathbb{V}_t(c_t)\bigr]
\;+\;
2\sum_{t=1}^T \mathbb{E}\bigl[\beta_t(c_t)\bigr].
\label{eq:ctx-rvu-final}
\end{align}
The potential term telescopes under standard regularity conditions on
$\Phi(\cdot,c)$, giving an $O(\Phi/\eta)$ contribution. Choosing $\eta$ in the
usual RVU manner and substituting the bound on $\sum_t \mathbb{V}_t(c_t)$ yields
the claimed regret rate in Proposition~5 (up to constants), with an additional
additive term governed by the cumulative context-dependent prediction radius
$\sum_t \beta_t(c_t)$.
\end{assumption}

We briefly outline how the preceding contextual analysis can be extended when
$\mathcal{C}$ is infinite (e.g., continuous).

\paragraph{Why the finite-context argument breaks.}
In the finite case, one can often take a union bound over all contexts to make
\eqref{eq:ctx-error} hold uniformly. When $\mathcal{C}$ is infinite, such a
uniform bound is generally impossible without additional structure.

\paragraph{Covering-number approach.}
Assume $(\mathcal{C},d)$ is a metric space and the prediction model is
Lipschitz in $c$:
\begin{equation}
|\hat r_t(a,c)-\hat r_t(a,c')|\le L\,d(c,c'), \qquad
|r_t(a,c)-r_t(a,c')|\le L_r\,d(c,c').
\end{equation}
Let $\mathcal{N}(\mathcal{C},d,\epsilon)$ be the covering number. One can prove
a high-probability version of \eqref{eq:ctx-error} on an $\epsilon$-net and then
transfer it to all contexts using Lipschitzness, leading to an error radius of
the form
\begin{equation}
\beta_t(c)
\;\lesssim\;
\sqrt{\frac{\log \mathcal{N}(\mathcal{C},d,\epsilon) + \log(1/\delta)}{n_t}}
\;+\; (L+L_r)\epsilon,
\end{equation}
where $n_t$ is an effective sample size (or information gain term, depending on
the estimator). Plugging this $\beta_t(\cdot)$ into
\eqref{eq:ctx-rvu-final} yields regret bounds with an additional complexity term
depending on $\log \mathcal{N}(\mathcal{C},d,\epsilon)$.

\paragraph{Function approximation view.}
Alternatively, one may model the context dependence via a parametric class
$\hat r_t(a,c)=\langle \theta_t, \phi(a,c)\rangle$ (linear) or a kernel / neural
predictor, replacing covering numbers with an effective dimension (e.g.,
elliptical potential / information gain). The resulting regret bound takes the
same form as \eqref{eq:ctx-rvu-final}, with $\sum_t \beta_t(c_t)$ controlled by
the complexity of the function class.

\paragraph{Takeaway.}
In infinite-context settings, the analysis typically reduces to controlling the context-conditional prediction radius $\beta_t(c)$ via capacity measures such as covering numbers or effective dimension; the rest of the RVU-style argument remains unchanged.

\section{Experimental Details}
\label{app:exp}
We provide detailed experimental setup, hyperparameters, and additional results.

\subsection*{Heads-Up No-Limit Texas Hold'em (1v1 HU-NLHE)}

In addition to 6-player NLHE, we evaluate various LLMs against the Slumbot (Brown \& Sandholm, 2017), a near-GTO bot based on CFR approximations. This provides a controlled benchmark to assess individual model capabilities in a simplified 2-player setting.

\paragraph{Experimental Setup.}
Each model plays 100 hands against Slumbot. Actions are extracted from the model's output via the \texttt{\textbackslash boxed\{\}} pattern. If extraction fails or the output exceeds 4096 tokens, the model defaults to \texttt{CHECK\_CALL}.

\begin{table}[!htbp]
  \centering
  \caption{Performance Comparison of AI Models in Poker}
  \scriptsize
  \setlength{\tabcolsep}{3pt}
  \renewcommand{\arraystretch}{1.05}
  \resizebox{\columnwidth}{!}{%
  \begin{tabular}{lcccccc}
    \toprule
    \textbf{Model} & \textbf{Avg. Return} & \textbf{vs. Slumbot} & \textbf{Better (\%)} & \textbf{Worse (\%)} & \textbf{Same (\%)} & \textbf{Action Extract (S/F)} \\
                  & \textbf{(BB/h)}       & \textbf{(BB/h)}      &                      &                     &                    &                               \\
    \midrule
    Qwen3-0.6B          & $-5.725$ & $-1.795$ & 44.00 & 52.00 & 4.00 & 280/26 \\
    Qwen3-1.7B          & $-3.520$ & $+2.785$ & 58.00 & 41.00 & 1.00 & 249/37 \\
    Qwen3-4B            & $+1.535$ & $+1.360$ & 47.00 & 53.00 & 0.00 & 179/79 \\
    Qwen3-4B-Thinking   & $+0.105$ & $-0.905$ & 44.00 & 54.00 & 2.00 & 107/125 \\
    Qwen3-4B-Instruct   & $-2.505$ & $+2.135$ & 52.00 & 46.00 & 2.00 & 277/0 \\
    GPT-5               & $-0.730$ & $+0.120$ & 47.00 & 51.00 & 2.00 & 281/24 \\
    GPT-5-mini          & $+2.355$ & $+2.175$ & 44.00 & 56.00 & 0.00 & 257/18 \\
    Claude 4            & $-0.065$ & $+1.010$ & 48.00 & 51.00 & 1.00 & 145/0 \\
    Claude 3.5          & $+0.260$ & $+0.280$ & 51.00 & 48.00 & 1.00 & 186/0 \\
    \bottomrule
  \end{tabular}%
  }
  \label{tab:poker_model_performance}
\end{table}

\begin{figure}[h]
    \centering
    \includegraphics[width=\textwidth]{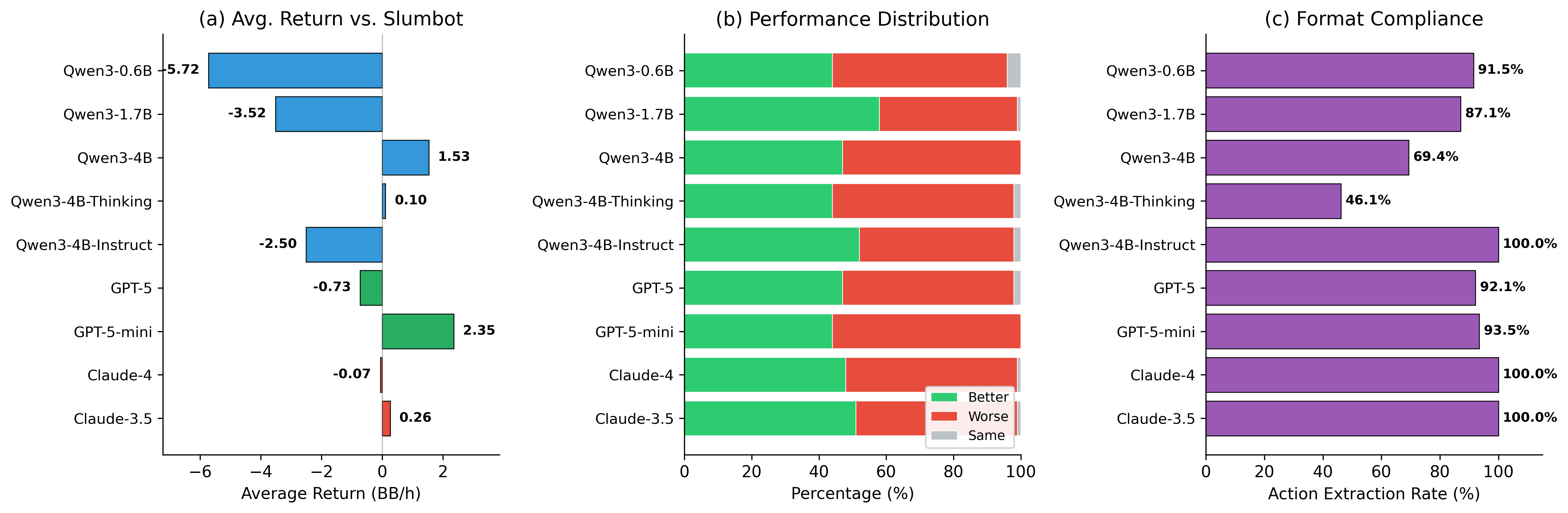}
    \caption{6-player NLHE results visualization. (a) LLM return comparison across models. (b) Heatmap showing per-position returns, with P6 as the LLM position.}
    \label{fig:poker_6player}
\end{figure}


\subsection{Details for Ablation Study}
\label{AS}
We conduct a set of ablation studies in 6-player NLHE to disentangle the contributions of different components in ISO-LLM, including supervised fine-tuning (SFT), strategic reward modeling (SRM), and group relative policy optimization (GRPO).

Table~\ref{tab:ablation_6player} reports the ablation results on long-term return (LTR) and win rate. Removing any single component leads to a clear degradation in LTR, despite relatively small changes in win rate. In particular, replacing the strategic reward model with a simple win/loss signal results in a $9.0$ BB/100 drop in LTR, highlighting the importance of delayed-value supervision. Removing GRPO further reduces LTR, indicating that optimization beyond reward modeling is necessary. Finally, training the policy from scratch without SFT yields negative LTR, demonstrating that a strong behavioral prior is critical for stable long-horizon learning.

\begin{table}[!htbp]
  \centering
  \caption{Ablation study in 6-player NLHE. We report LTR (BB/100), Win Rate (\%), and the change in LTR relative to the full ISO-LLM model.}
  \label{tab:ablation_6player}
  \begin{tabular}{lccc}
    \toprule
    \textbf{Variant} & \textbf{LTR (BB/100)} & \textbf{Win Rate (\%)} & \textbf{$\Delta$LTR} \\
    \midrule
    \textbf{ISO-LLM (full)}          & $+15.8$ & 22 & --- \\
    w/o SRM (win/loss reward)        & $+6.8$  & 21 & $-9.0$ \\
    w/o GRPO (SFT+SRM only)          & $+9.2$  & 19 & $-6.6$ \\
    w/o SFT (from scratch)           & $-2.1$  & 12 & $-17.9$ \\
    \bottomrule
  \end{tabular}
\end{table}

While the above ablation isolates the contribution of individual training components, it does not yet disentangle the algorithmic novelty of ISO from improvements due to reward modeling or RL optimization alone. To this end, we further analyze the role of \emph{context-conditioned learning} in ISO-GRPO.

Table~\ref{tab:ablation-context-components} compares ISO-GRPO with several GRPO-based variants that differ in how contextual predictions are used. Simply providing predicted context as an additional input feature yields only marginal improvements over vanilla GRPO. In contrast, explicitly separating learners by context and using predictions for both routing and per-context updates substantially improves LTR while simultaneously reducing exploitability. An oracle variant with access to the true context $Z_t$ provides an upper bound, confirming that the remaining gap is primarily due to prediction error rather than optimization limitations.

Taken together, these results indicate that the performance gains of ISO-LLM do not arise solely from stronger reward signals or improved RL optimization. Instead, they crucially depend on the proposed prediction-aware learning rule, which leverages context separation and routing to better handle long-horizon strategic non-stationarity.

\begin{table}[h]
  \centering
  \caption{Ablation study of context-conditioned learning components}
  \label{tab:ablation-context-components}
  \footnotesize
  \setlength{\tabcolsep}{3pt} 
  \renewcommand{\arraystretch}{1.15}
  \resizebox{\columnwidth}{!}{%
  \begin{tabular}{lccccc}
    \toprule
    \textbf{Method} & \textbf{Context separation} & \textbf{Prediction used for routing} & \textbf{LTR (BB/100)} & \textbf{Exploitability} \\
    \midrule
    GRPO (no-context) & $\times$ & $\times$ & 2.7 & 7.6 \\
    GRPO + Pred-as-feature & $\times$ & $\surd$ (feature only) & 3.1 & 7.1 \\
    \textbf{ISO-GRPO (ours)} & $\surd$ & $\surd$ (routing + per-ctx update) & \textbf{6.3} & \textbf{4.9} \\
    ISO-GRPO (oracle routing) & $\surd$ & $\surd$ (true $Z_t$) & 6.8 & 4.7 \\
    \bottomrule
  \end{tabular}%
  }
\end{table}

\subsection{Qualitative Examples of Emergent Long-Horizon Behavior}
\label{subsec:qualitative}

\paragraph{Example 1: River bluff-raise (NLHE).}
\begin{itemize}
    \item \textbf{Situation:} ISO-LLM reaches the river with a missed flush draw; pot $=50$BB; opponent bets 25BB.
    \item \textbf{Myopic action:} Fold (no showdown value).
    \item \textbf{ISO action:} Raise to 75BB (bluff).
    \item \textbf{Interpretation:} SRM assigns high long-horizon value given pot geometry, board texture, and opponent line. Even when the bluff fails, it may shape future opponent responses (table image), yielding delayed strategic benefit.
\end{itemize}

\paragraph{Example 2: Sacrifice play (Pok\'{e}mon).}
\begin{itemize}
    \item \textbf{Situation:} A weakened Alakazam faces a full-health Snorlax; switching to Gengar is safe.
    \item \textbf{Myopic action:} Switch to preserve Alakazam.
    \item \textbf{ISO action:} Use Psychic with Alakazam, accepting the KO.
    \item \textbf{Interpretation:} SRM captures that sacrificing Alakazam can enable a free Gengar entry and a higher-probability sweep line, increasing long-horizon win chance.
\end{itemize}

\section{Demonstration of Pokemon Battle}

\subsection{Case Studies: Long-term Strategic Planning}

We present three representative case studies from the trained ISO policy that illustrate emergent long-term planning behaviors. In each case, the agent deliberately sacrifices short-term advantage (e.g., HP, tempo, or a Pokémon) to secure a more favorable long-horizon outcome.

\paragraph{Example 1: Strategic HP Trading for Type Advantage.}
\label{ex:hp_trading}

\begin{quote}

\textbf{Context.} The agent (Player 1) faces a disadvantageous matchup in which the opponent’s Snorlax represents the primary late-game win condition.

\begin{center}
\small
\begin{tabular}{cccc}
\toprule
\textbf{Turn} & \textbf{Our State} & \textbf{Action} & \textbf{Opponent State} \\
\midrule
22 & Snorlax 242/523 HP & Switch to Chansey & Snorlax 50\% \\
23 & Chansey 508/703 HP & Thunder Wave & Alakazam 100\% \\
24 & Chansey 703/703 HP & Soft-Boiled & Alakazam 100\% \\
$\vdots$ & \multicolumn{3}{c}{\textit{(Strategic stalling over 20+ turns)}} \\
45 & Tauros 239/353 HP & Body Slam & Snorlax 49\% \\
45 & Tauros \textbf{Fainted} & -- & Snorlax \textbf{Fainted} \\
\bottomrule
\end{tabular}
\end{center}

\textbf{Analysis.} The agent intentionally accepts a one-for-one trade by sacrificing Tauros to eliminate the opponent’s Snorlax. Although this decision incurs an immediate HP loss (353$\rightarrow$0), it removes the opponent’s primary win condition, yielding a decisive long-term strategic advantage.
\end{quote}

\paragraph{Example 2 : Delayed Sweep Setup via Sacrificial Status Spreading}
\label{ex:delayed_sweep}

\begin{quote}

\textbf{Context.} The agent maintains a weakened Exeggutor to gradually cripple the opponent’s defensive core, rather than pursuing immediate damage.

\begin{center}
\small
\begin{tabular}{cccc}
\toprule
\textbf{Turn} & \textbf{Our HP} & \textbf{Action / Role} & \textbf{Opponent State} \\
\midrule
27 & Exeggutor 311/393 & Stun Spore & Chansey 52\% \\
28--52 & Fluctuating & Status spreading & Gradual decline \\
55 & Exeggutor 211/393 & Psychic & Chansey 40\% \\
56--60 & Switch cycle & Recovery timing & Starmie 100\% \\
78 & Exeggutor \textbf{Fainted} & -- & Rhydon 72\% (crippled) \\
\bottomrule
\end{tabular}
\end{center}

\textbf{Final Result.} Win after 101 turns, with the opponent’s team fully swept.

\textbf{Analysis.} Instead of maximizing immediate damage, the agent repurposes Exeggutor as a long-horizon enabler by spreading paralysis and sleep. Although Exeggutor is eventually sacrificed, this delayed setup enables a decisive sweep by the remaining team, reversing the HP disadvantage observed at turn 78.
\end{quote}

\begin{example}[Predictive Counter-Switching under HP Disadvantage]
\label{ex:counter_switch}
\textbf{Context.} The agent performs a sequence of predictive switches to preserve type advantage, despite incurring substantial short-term HP loss.

\begin{center}
\small
\begin{tabular}{ccccc}
\toprule
\textbf{Turn} & \textbf{Our Pokémon} & \textbf{HP} & \textbf{Opponent Pokémon} & \textbf{Opp. HP} \\
\midrule
3 & Starmie & 537/703 & Gengar & 100\% \\
4 & \textbf{Switch$\rightarrow$Golem} & 192/363 & Tauros (switch) & 100\% \\
5 & \textbf{Switch$\rightarrow$Moltres} & 146/383 & Tauros & 100\% \\
6 & \textbf{Switch$\rightarrow$Snorlax} & 292/523 & Tauros & 100\% \\
7 & Snorlax & 292/523 & Tauros & 39\% \\
8 & Snorlax & 73/523 & Tauros & 9\% \\
\bottomrule
\end{tabular}
\end{center}

\textbf{HP Accounting.}
\begin{itemize}
  \item \textbf{Turns 3--6:} Cumulative HP loss $\approx 634$ across three Pokémon.
  \item \textbf{Turns 7--8:} Opponent Tauros reduced from 100\% to 9\%.
  \item \textbf{Net Effect:} Removal of the opponent’s primary sweeper.
\end{itemize}

\textbf{Final Outcome.} Victory at turn 43 with one Pokémon remaining.

\textbf{Analysis.} The agent tolerates a large distributed HP loss in exchange for maintaining favorable matchups and eliminating Tauros. This behavior reflects forward-looking evaluation of team-level outcomes rather than short-term survivability.
\end{example}

\subsection{Quantitative Summary}

\begin{table}[!htbp]
\centering
\caption{Long-term Planning Metrics from Case Studies}
\begin{tabular}{|l|c|c|c|}
\hline
\textbf{Metric} & \textbf{Case 1} & \textbf{Case 2} & \textbf{Case 3} \\
\hline
Total Battle Length (turns) & 85 & 101 & 43 \\
Initial HP Disadvantage Turns & 15 & 28 & 6 \\
Strategic Sacrifice Count & 1 & 2 & 3 \\
HP Recovery Before Victory & +450 & +890 & +280 \\
Final Victory Margin (Pokémon) & 3-0 & 1-0 & 1-0 \\
\hline
\end{tabular}
\label{tab:planning_metrics}
\end{table}

These examples demonstrate that our RL-trained policy has learned to:
\begin{enumerate}
\item Accept short-term HP losses for positional advantage
\item Execute multi-turn setup strategies (status spreading, chip damage)
\item Predict opponent switches and counter appropriately
\item Sacrifice individual Pokémon for team-level victory conditions
\end{enumerate}

This behavior aligns with the ``delayed gratification'' characteristic of effective long-term planning, where the agent prioritizes eventual victory over immediate reward maximization.

\newpage
\section{Real Cases Analysis in Texas Hold'em}
\subsection{Case 1: Strategic Folding of Extremely Weak Hands in Early Position}
\textbf{Input:}

\begin{table*}[!ht]
\centering
\caption{One case of weak hands in early position.}
\label{ex:11}

\begin{mdframed}[
  linewidth=0.5pt,
  linecolor=black,
  backgroundcolor=black!5,
  innertopmargin=4pt,
  innerbottommargin=4pt,
  innerleftmargin=6pt,
  innerrightmargin=6pt
]
\begin{lstlisting}[
  basicstyle=\footnotesize\ttfamily,  % 更小的字体
  breaklines=true,                   % 自动换行
  breakatwhitespace=false,           % 允许在任何位置换行
  prebreak=\mbox{\textcolor{red}{\tiny$\hookrightarrow$}\space},  % 换行指示符
  postbreak=\mbox{\textcolor{red}{\tiny$\hookrightarrow$}\space}, % 换行指示符
  columns=fullflexible,              % 完全灵活的列
  showstringspaces=false,            % 不显示字符串中的空格
  xleftmargin=0pt,                   % 去除左边空白
  xrightmargin=0pt,                  % 去除右边空白
  aboveskip=0pt,                     % 去除上方空白
  belowskip=0pt                      % 去除下方空白
]
History of the game: [ ]
Hand Cards: ["7D", "2S"]
Board Cards: [ ]
Number of Players: 6
Hero Position: 3
Button Position: 0
All Chips: [0, 1, 2, 0, 0, 0]
Pot: 3
Stakes: [500, 499, 498, 500, 500, 500]
\end{lstlisting}
\end{mdframed}

\end{table*}

\textbf{Solution Comparison:}
\begin{table}[!htbp]
    \centering
        \caption{Weak Hand Handling Comparison}
    \begin{tabular}{|p{0.2\textwidth}|p{0.35\textwidth}|p{0.35\textwidth}|}
        \hline
        \textbf{Model} & \textbf{ISO Model} & \textbf{Baseline Model} \\
        \hline
         \textbf{Explanation} & Hero holds 72 offsuit in UTG - the worst starting hand in poker. With 3 players behind us, playing this hand has negative expected value. Even occasional blind steals cannot compensate for playing a dominated hand out of position. Fold and preserve chips for profitable spots. & Hero is first to act with 72 offsuit. A standard raise can win the blinds immediately if all opponents fold. Taking initiative preflop establishes table presence and puts pressure on other players. \\
        \hline
         \textbf{Action} & FOLD & RAISE\_POT\\
        \hline
    \end{tabular}
\end{table}

\paragraph{Note:} The baseline model's "aggressive by default" pattern appears profitable short-term but hemorrhages chips over hundreds of hands.
\newpage

\subsection{Case 2: Monster Hand Slow-Play Trap}
\textbf{Input:}

\begin{table*}[!ht]
\centering
\caption{One case of monster hand handling.}
\label{ex:11}

\begin{mdframed}[
  linewidth=0.5pt,
  linecolor=black,
  backgroundcolor=black!5,
  innertopmargin=4pt,
  innerbottommargin=4pt,
  innerleftmargin=6pt,
  innerrightmargin=6pt
]
\begin{lstlisting}[
  basicstyle=\footnotesize\ttfamily,  % 更小的字体
  breaklines=true,                   % 自动换行
  breakatwhitespace=false,           % 允许在任何位置换行
  prebreak=\mbox{\textcolor{red}{\tiny$\hookrightarrow$}\space},  % 换行指示符
  postbreak=\mbox{\textcolor{red}{\tiny$\hookrightarrow$}\space}, % 换行指示符
  columns=fullflexible,              % 完全灵活的列
  showstringspaces=false,            % 不显示字符串中的空格
  xleftmargin=0pt,                   % 去除左边空白
  xrightmargin=0pt,                  % 去除右边空白
  aboveskip=0pt,                     % 去除上方空白
  belowskip=0pt                      % 去除下方空白
]
History of the game:
[
  {"player number": 3, "action": "RAISE_POT", "amount": 6, "game stage": "Pre-Flop"},
  {"player number": 4, "action": "FOLD", "amount": 0, "game stage": "Pre-Flop"},
  {"player number": 5, "action": "FOLD", "amount": 0, "game stage": "Pre-Flop"},
  {"player number": 0, "action": "FOLD", "amount": 0, "game stage": "Pre-Flop"},
  {"player number": 1, "action": "FOLD", "amount": 0, "game stage": "Pre-Flop"},
  {"player number": 2, "action": "CHECK_CALL", "amount": 4, "game stage": "Pre-Flop"},
  {"player number": 2, "player role": "BB", "action": "CHECK", "amount": 0, "game stage": "Flop"}
]

Hand Cards: ["AS", "AD"]
Board Cards: ["AH", "7C", "2D"]
Number of Players: 6
Hero Position: 3
Button Position: 0
All Chips: [0, 1, 6, 6, 0, 0]
Pot: 13
Stakes: [500, 499, 494, 494, 500, 500]
\end{lstlisting}
\end{mdframed}

\end{table*}

\textbf{Solution Comparison:}
\begin{table}[!htbp]
    \centering
        \caption{Monster Hand Handling Comparison}
    \begin{tabular}{|p{0.2\textwidth}|p{0.35\textwidth}|p{0.35\textwidth}|}
        \hline
        \textbf{Model} & \textbf{ISO Model} & \textbf{Baseline Model} \\
        \hline
         \textbf{Explanation} & Hero flopped top set with AA on a dry A72 rainbow board. Villain checked to us. Betting now folds out most of their range. By checking, we let villain catch up or bluff the turn, allowing us to extract more chips later rather than winning a small pot now. & Hero flopped top set with AA on a A72 rainbow board. We have an extremely strong hand and should bet for value immediately. A pot-sized bet protects against draws and builds the pot while we're ahead. \\
        \hline
         \textbf{Action} & CHECK & RAISE\_POT\\
        \hline
    \end{tabular}
\end{table}

\paragraph{Note:}The baseline model's "bet every monster hand" pattern wins small pots immediately but misses opportunities to extract maximum value from weaker hands that could improve on later streets.
\newpage

\subsection{The Wise Fold After Recognising an Opponent's Strong Hand Signal}
\textbf{Input}
\begin{table*}[!ht]
\centering
\caption{One case of recognizing an opponent's strong hand signal.}
\label{ex:strong-signal}

\begin{mdframed}[
  linewidth=0.5pt,
  linecolor=black,
  backgroundcolor=black!5,
  innertopmargin=4pt,
  innerbottommargin=4pt,
  innerleftmargin=6pt,
  innerrightmargin=6pt
]
\begin{lstlisting}[
  basicstyle=\footnotesize\ttfamily,  % 更小的字体
  breaklines=true,                   % 自动换行
  breakatwhitespace=false,           % 允许在任何位置换行
  prebreak=\mbox{\textcolor{red}{\tiny$\hookrightarrow$}\space},  % 换行指示符
  postbreak=\mbox{\textcolor{red}{\tiny$\hookrightarrow$}\space}, % 换行指示符
  columns=fullflexible,              % 完全灵活的列
  showstringspaces=false,            % 不显示字符串中的空格
  xleftmargin=0pt,                   % 去除左边空白
  xrightmargin=0pt,                  % 去除右边空白
  aboveskip=0pt,                     % 去除上方空白
  belowskip=0pt                      % 去除下方空白
]
History of the game:
[
  {"player number": 3, "action": "RAISE_POT", "amount": 6, "game stage": "Pre-Flop"},
  {"player number": 4, "action": "FOLD", "amount": 0, "game stage": "Pre-Flop"},
  {"player number": 5, "action": "FOLD", "amount": 0, "game stage": "Pre-Flop"},
  {"player number": 0, "action": "CHECK_CALL", "amount": 6, "game stage": "Pre-Flop"},
  {"player number": 1, "action": "FOLD", "amount": 0, "game stage": "Pre-Flop"},
  {"player number": 2, "action": "FOLD", "amount": 0, "game stage": "Pre-Flop"},
  {"player number": 3, "action": "RAISE_HALF_POT", "amount": 7, "game stage": "Flop"},
  {"player number": 0, "action": "RAISE_POT", "amount": 21, "game stage": "Flop"},
  {"player number": 3, "action": "ALL_IN", "amount": 466, "game stage": "Flop"}
]

Hand Cards: ["JH", "JD"]
Board Cards: ["KC", "KH", "4S"]
Number of Players: 6
Hero Position: 0
Button Position: 0
All Chips: [27, 1, 2, 472, 0, 0]
Pot: 502
Stakes: [473, 499, 498, 28, 500, 500]
\end{lstlisting}
\end{mdframed}

\end{table*}

\textbf{Solution Comparison:}
\begin{table}[!htbp]
    \centering
    \caption{Monster Hand Handling Comparison}
    \begin{tabular}{|p{0.2\textwidth}|p{0.35\textwidth}|p{0.35\textwidth}|}
        \hline
        \textbf{Model} & \textbf{ISO Model} & \textbf{Baseline Model} \\
        \hline
         \textbf{Explanation} & Hero holds JJ facing an all-in on a K-K-4 board. Villain opened UTG, c-bet, and shoved after our raise - this extreme aggression on a paired King board indicates trips or a full house. Our equity against this range is under 0.05. Folding saves 445 chips despite holding a 'good hand'. & Hero holds JJ facing an all-in. Pocket Jacks is a premium hand and we already invested 27 chips. The pot odds are approximately 2:1, which means we only need 0.33 equity to call profitably. JJ should have enough equity against villain's range. \\
        \hline
         \textbf{Action} & FOLD & CHECK\\
        \hline
    \end{tabular}
    
\end{table}

\paragraph{Note:}This comparison illustrates how the ISO Model’s more precise equity assessment—accounting for the villain’s range and board texture—leads to a more profitable, risk-aware decision, whereas the Baseline Model relies on generalized assumptions about hand strength and pot odds.
\newpage

\section{Environment}
\label{app:env}
\subsection{LLM Poker Environment}
We constructed a multi-agent poker environment, and the code will be released upon acceptance.  We evaluate our method in the heterogeneous multi-agent poker environment designed for long-horizon decision-making under imperfect information.
The environment supports mixed populations of agents, including rule-based policies, conventional RL agents, and LLM-based agents, all interacting through a unified action interface.

\begin{figure}[h]
    \centering
    \includegraphics[width=0.9\linewidth]{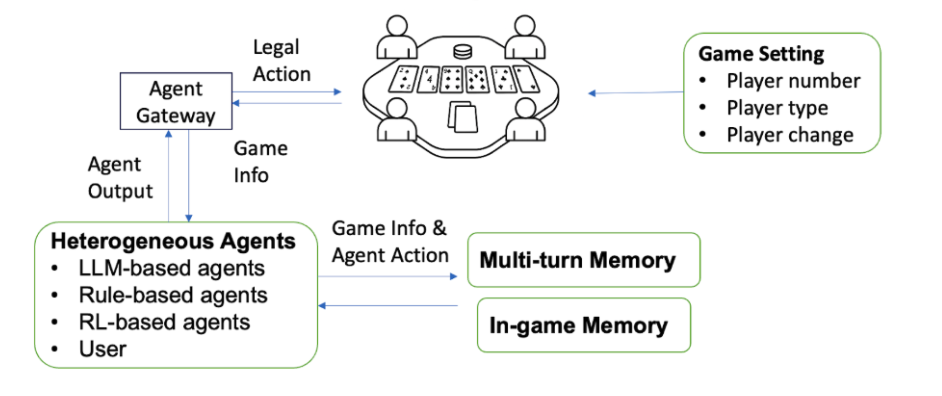}
    \caption{Overview of the LLM Poker environment. The environment supports heterogeneous agents, a unified action gateway, and memory mechanisms for long-horizon strategic interaction.}
    \label{fig:poker-env}
\end{figure}

At the beginning of each hand, a game configuration specifies the number of players, blind structure, stack sizes, and betting rules.
Players proceed through standard poker stages (pre-flop, flop, turn, river), with action legality enforced by the environment.

\subsection{Observations, Actions, and Memory}
Each agent observes its private cards, public cards, pot size, stack sizes, player positions, and the full betting history.
Actions follow standard no-limit poker operations (check, call, bet, raise, fold), with bet sizing constrained by game rules.

To support long-horizon analysis, the environment maintains both within-hand memory (action history and intermediate states) and cross-hand memory (aggregated behavioral statistics).
For prediction-aware evaluation, we additionally log a discrete context label $Z_t$ at each decision point, which is not revealed to agents before acting but is used to measure prediction accuracy and misprediction counts in our experiments.

\bibliographystyle{plainnat}
\bibliography{example_paper}

\end{document}